\documentclass[a4paper,fleqn]{cas-sc}

\usepackage[authoryear,longnamesfirst]{natbib}
\usepackage{tikz}
\usepackage{graphicx}

\def\tsc#1{\csdef{#1}{\textsc{\lowercase{#1}}\xspace}}
\tsc{WGM}
\tsc{QE}

\newcommand*\ddiff{\mathop{}\!\mathrm{d}}

\newcommand{\dd}[2] {\ensuremath{\frac{\partial {#1}}{\partial {#2}}}}
\newcommand{\DD}[2] {\ensuremath{\frac{\ddiff {#1}}{\ddiff {#2}}}}

\newcommand{\LRp}[1]{\left( #1 \right)}

\newcommand{\norm}[1]{\left\lVert#1\right\rVert}

\newcommand{\xb}{{\bf x}}
\newcommand{\fb}{{\bf f}}

\begin{document}
\let\WriteBookmarks\relax
\def\floatpagepagefraction{1}
\def\textpagefraction{.001}

\shorttitle{Knowledge-Integrated Neural Modeling for the SMIB Systems}

\shortauthors{Kang et~al.}

\title [mode = title]{Knowledge Integration in Differentiable Models: A Comparative Study of Data-Driven, Soft-Constrained, and Hard-Constrained Paradigms for Identification and Control of the Single Machine Infinite Bus System}

\tnotemark[1]

\tnotetext[1]{This work was supported by Korea University Grants (No. K2411041, K2414071, K2425851, K2514791).}

\author[1]{Shinhoo Kang}[
    orcid=0000-0003-4649-9727]
\cormark[1]
\ead{shinkang@korea.ac.kr}
\credit{Original draft, Conceptualization, Methodology, Formal analysis, Visualization, Software}
\affiliation[1]{organization={Department of Computer Science and Software Engineering, Korea University},
            addressline={2511 Sejong-Ro},
            city={Sejong},
            postcode={30019},
            country={Republic of Korea}}

\author[1]{Sangwook Kim}
\ead{sw2067@korea.ac.kr}
\credit{Review and editing, Visualization, Software}

\author[2]{Sehyun Yun}
\ead{sehyuny@hyundai.com}
\credit{Review and editing}
\affiliation[2]{organization={Department of Autonomous Driving at Hyundai Motor Company},
            city={Seoul},
            postcode={06797},
            country={Republic of Korea}}

\cortext[1]{Corresponding author}

\begin{abstract}
Integrating domain knowledge into neural networks is a central challenge in scientific machine learning. Three paradigms have emerged---data-driven (Neural Ordinary Differential Equations, NODEs), soft-constrained (Physics-Informed Neural Networks, PINNs), and hard-constrained (Differentiable Programming, DP)---each encoding physical knowledge at different levels of structural commitment. However, how these strategies impact not only predictive accuracy but also downstream tasks such as control synthesis remains insufficiently understood.
This paper presents a comparative study of NODEs, PINNs, and DP for dynamical system modeling, using the Single Machine Infinite Bus power system as a benchmark. We evaluate these paradigms across three tasks: trajectory prediction, parameter identification, and Linear Quadratic Regulator control synthesis.
Our results yield three principal findings. First, knowledge representation determines generalization: NODE, which learns the system operator, enables robust extrapolation, whereas PINN, which approximates a solution map, restricts generalization to the training horizon. Second, hard-constrained formulations (DP) reduce learning to a low-dimensional physical parameter space, achieving faster and more reliable convergence than soft-constrained approaches. Third, knowledge fidelity propagates to control performance: DP produces controllers that closely match those obtained from true system parameters, while NODE provides a viable data-driven alternative by recovering control-relevant Jacobians with $3-4\%$ relative error and yielding LQR gains within $0.36\%$ of the ground truth.
Based on these findings, we propose a practical decision framework for selecting knowledge integration strategies in neural modeling of dynamical systems.
\end{abstract}


\begin{keywords}
Knowledge Integration \sep
Physics-Informed Neural Networks \sep
Neural Ordinary Differential Equations \sep
Differentiable Programming \sep
Scientific Machine Learning \sep
Inductive Bias
\end{keywords}

\maketitle

\section{Introduction}

Integrating domain knowledge into data-driven models is a central challenge in artificial intelligence.
In scientific and engineering applications, physical laws provide powerful prior knowledge 
that can dramatically improve model accuracy, generalization, and interpretability \citep{karniadakis2021physics,von2021informed}. However, a fundamental question remains insufficiently understood:
\emph{how does the choice of knowledge encoding strategy affect the behavior and reliability of learned models?}

This question is critical in dynamical systems, 
where learned models are not only expected to predict observed trajectories 
but also to support downstream tasks such as parameter identification, stability analysis, and control design.
In these settings, different strategies for incorporating prior knowledge may lead to fundamentally different trade-offs between predictive accuracy, physical consistency, and robustness under measurement noise. 

In recent years, three distinct paradigms have emerged in scientific machine learning, 
each representing a different approach to knowledge integration. 
\emph{Data-driven} approaches, such as Neural Ordinary Differential Equations (NODEs) \citep{chen2018neural},
learn the system dynamics directly from data without explicit physics encoding,
relying on the structural inductive bias of continuous-time modeling. 
\emph{Soft-constrained} approaches, exemplified by Physics-Informed Neural Networks (PINNs) \citep{raissi2019physics},
embed governing equations into the training loss as penalty terms, encouraging but not strictly enforcing physical consistency.
\emph{Hard-constrained} approaches based on Differentiable Programming (DP) \citep{rackauckas2020universal},
embed known physical laws directly into the computational graph, restricting learning to unknown physics while preserving the governing equations. 
These three paradigms span a knowledge integration spectrum from minimal to maximal structural commitment,
yet their comparative implications remain insufficiently explored within a unified framework.

Power systems provide a compelling testbed for this investigation.
The increasing penetration of renewable energy sources and inverter-based resources has led to low-inertia grid conditions,
where system dynamics become highly sensitive to disturbances and control actions \citep{milano2018foundations,ratnam2020future}.
In such systems, models should not only provide accurate trajectories but also support parameter estimation and control design \citep{zhong2010synchronverters,elenga2024challenges}.
At the same time, modern Internet-of-Things sensing infrastructures enable high-resolution measurements \citep{saleem2019internet,abir2021iot}, 
while practical challenges such as measurement noise persists \citep{zhang2014failure}.
These characteristics make power system dynamics an ideal setting for evaluating how different knowledge integration strategies perform under realistic conditions.

Despite the architectural elegance of these paradigms,
existing studies have largely evaluated these paradigms in isolation or through task-specific comparisons,
 with limited effort to systematically assess their relative strengths and weaknesses across prediction,
  identification, and control within a unified framework.
In particular, while trajectory fitting is often used as the primary evaluation criterion, 
it does not necessarily guarantee that the learned models provide physically meaningful system sensitivities 
required for downstream tasks such as Linear Quadratic Regulator (LQR) design. 
As a result, it remains unclear how different knowledge integration paradigms affect not only prediction,
 but also identification and control performance in a consistent and systematic manner.

To bridge this gap,
this paper presents a unified comparative study of three knowledge integration paradigms---NODE (data-driven), PINN (soft-constrained), and DP (hard-constrained)---using the Single Machine Infinite Bus (SMIB) system as a canonical benchmark. 
We evaluate these approaches using three tasks: trajectory forecasting, physical parameter identification, and control-relevant system consistency under noisy conditions. 
Through this study, we aim to elucidate how the choice of knowledge encoding strategy shapes the trade-offs between accuracy, robustness, and control reliability in neural modeling of dynamical systems.

The main contributions are summarized as follows:

\begin{itemize}
    \item
We establish a unified comparative framework for evaluating 
three representative knowledge integration paradigms—NODE, PINN, and DP—in neural modeling of dynamical systems.
    \item
We provide a systematic evaluation of these approaches across three key tasks:
     trajectory forecasting, physical parameter identification, and control-oriented consistency under noisy conditions.
    \item
We demonstrate that different knowledge encoding strategies induce fundamentally different trade-offs
 between predictive accuracy, inverse problem convergence, and control reliability.
    \item
We offer practical insights into the strengths and limitations of soft-constrained, data-driven, and hard-constrained approaches,
 providing guidance for selecting appropriate modeling strategies in power system applications.
\end{itemize}

The rest of this paper is organized as follows.
Section~\ref{section:related-work} reviews related work on knowledge-informed machine learning, covering NODEs, PINNs, DP, and classical approaches.
Section~\ref{section:physical-model} presents the mathematical formulation of the SMIB system dynamics and defines the problem of parameter identification. Section~\ref{section:differentiable-model} details the three differentiable modeling paradigms. Section~\ref{section:numerical-results} provides a comparative analysis of prediction, identification, and control performance. Finally, Section~\ref{section:conclusion} concludes the paper.

\section{Related Work}
\label{section:related-work}

This section reviews the literature on integrating domain knowledge into machine learning models for dynamical systems. We organize the discussion along the knowledge-integration spectrum---from general taxonomies of informed machine learning, the three paradigms considered in this work (NODEs, PINNs, and DP), and finally to classical non-machine learning approaches for system identification.

\subsection{Knowledge-Informed Machine Learning}

The question of how to systematically incorporate domain knowledge into machine learning has attracted significant attention in the knowledge-based systems and artificial intelligence communities. 
\citet{von2021informed} proposed a comprehensive taxonomy of \emph{informed machine learning},
 categorizing knowledge integration along three dimensions: the source of knowledge (scientific knowledge, expert knowledge, or world knowledge), 
 its representation (algebraic equations, differential equations, logic rules, or simulation results), 
 and its integration point (training data, hypothesis set, learning algorithm, or final hypothesis). 

The three paradigms studied in this work---NODE, PINN, and DP---share the same knowledge source and representation: all three draw on physical laws expressed as differential equations. They differ in the integration point at which this knowledge is incorporated into the learning process. 
NODE incorporates the weakest form of such knowledge, assuming only that the system evolves according to a continuous-time ODE. PINN introduces physics-based loss terms that penalize violations of the governing equations. DP enforces the strongest form of knowledge integration by embedding the exact governing equations directly into the hypothesis space as a differentiable solver, ensuring that all candidate models are, by construction, physically consistent with the governing dynamics.
From this perspective, the three paradigms form a natural progression along the knowledge integration axis, ranging from structural priors (NODE), to soft residual-based constraints (PINN), to hard architectural constraints (DP).
 
This progression aligns closely with the notion of \emph{inductive bias} as an organizing principle for understanding knowledge integration in neural models \citep{goyalinductive2022}.
NODE imposes a minimal bias through continuous-time structure,
PINN imposes a soft bias via physics-based loss regularization,
and DP enforces a hard bias by embedding exact governing equations as differentiable computational layers. 
Our work contributes to this line of research by empirically demonstrating how the strength of inductive bias influences predictive performance and downstream control behavior.

\subsection{Neural ODEs}

Neural Ordinary Differential Equations (NODEs),
 proposed by \citet{chen2018neural}, parameterize the right-hand side of an ODE using a neural network, 
 with state trajectories obtained via numerical integration. This formulation provides a continuous-time and memory-efficient alternative to discrete residual networks and has become a cornerstone of latent dynamics modeling.
 
Several extensions have been proposed to address limitations of the original framework. 
\citet{dupontaugmented2019} introduced augmented NODEs that expand the state space to overcome topological constraints on learned flows. 
\citet{norcliffesecond2020} proposed second-order NODEs that directly model Newtonian dynamics, 
offering a physically motivated architecture for mechanical systems. 
\citet{rubanova2019latent} introduced latent ODEs, which combine NODEs with variational inference to learn continuous-time latent dynamics from irregularly sampled and partially observed data.

In the power systems context, NODEs have been applied to frequency and transient dynamics modeling. \citet{xiaofeasibility2023} demonstrated their feasibility for power system dynamic simulation and showed generalization across operating conditions. \citet{aslamipower2024} modeled transients under varying fault scenarios, while \citet{karacelebipower2024} explored NODE-based approaches for real-time dynamic security assessment.
While these studies demonstrate the flexibility of NODEs as data-driven surrogates, they do not systematically evaluate the fidelity of learned sensitivities for control. This gap forms a central focus of our work.

\subsection{Physics-Informed Neural Networks}

Physics-Informed Neural Networks (PINNs), introduced by \citet{raissi2019physics}, 
embed governing differential equations into the training loss as soft constraints.
By penalizing violations of physical laws at collocation points,
PINNs enable simultaneous data fitting and physics enforcement, even in the absence of labeled solution data.
This framework has been widely adopted across scientific and engineering domains.

However, several fundamental challenges have been identified.
\citet{wangwhen2021} showed that PINNs may fail to recover correct solutions due to imbalanced gradients between data and physics loss terms,
 and proposed adaptive weighting strategies to mitigate this issue.
  The difficulty of training PINNs for stiff or highly oscillatory systems has been further analyzed by \citet{krishnapriyancharacterizing2021},
   who demonstrated that the loss landscape can become highly ill-conditioned,
    leading to severe optimization challenges for certain classes of differential equations.

In the power systems domain, PINNs have been applied to state estimation and parameter identification. 
\citet{misyrisphysicsinformed2020} pioneered their use for transient stability analysis,
 demonstrating the ability to estimate dynamic states from limited measurements.
\citet{ngophysicsinformed2024} extended this framework to account for measurement noise and partial observability.
\citet{zoupinnsdriven2025} applied PINNs to parameter identification in multi-machine systems, 
while \citet{nadalphysicsinformed2025} explored gradient-enhanced formulations to improve sensitivity recovery.

Despite these advances, existing studies primarily focus on predictive accuracy, and applying PINNs to closed-loop control presents additional structural challenges. 
PINNs parameterize the solution map of a specific initial-boundary value problem, rather than the underlying vector field. Consequently, evaluating the system response under new control inputs or initial conditions may require retraining. Alternatively, one must learn an extended solution map whose input dimensionality scales with the number of control and state variables. This setting is known to make PINN training significantly more challenging due to loss-balancing and optimization pathologies. 
For this reason, our benchmark includes PINNs in the prediction and parameter-identification experiments but restricts the control study to modeling paradigms---NODE and DP---that directly expose the vector field and its sensitivities in a form compatible with control design.

 \subsection{Differentiable Programming as a Hybrid Physics-Neural Approach}

The term \emph{differentiable programming} (DP) broadly refers to the practice of making entire computational pipelines differentiable via automatic differentiation. 
In scientific machine learning, however, DP takes on a more specific meaning: it denotes a \emph{hybrid} modeling paradigm that couples physics-based simulation with gradient-based learning. 
Concretely, known governing equations---typically implemented as numerical ODE or PDE solvers---are embedded as differentiable layers within a computational graph. 
Unknown physical parameters are then optimized via backpropagation through the solver, analogous to training neural network weights. 
This hybrid formulation distinguishes DP from purely data-driven approaches, which learn dynamics directly from data, 
and from classical numerical methods, which lack learnable components. 
DP preserves the full physical structure while enabling efficient gradient-based optimization. 

The closely related framework of \emph{Universal Differential Equations} (UDEs) \citep{rackauckas2020universal} can be viewed as a specific realization of the DP paradigm, in which known differential equations are augmented with neural network components that learn unknown or misspecified terms.

Differentiable simulation has been successfully applied across scientific domains. 
\citet{hudifftaichi2020} developed DiffTaichi, enabling gradient-based optimization of complex physical systems. In power systems, \citet{hyett2024differentiable} demonstrated differentiable simulation for system optimization, while \citet{rosembergdifferentiable2025} explored its application to optimal power flow. 
Similar hybrid strategies have also been adopted in fluid mechanics \citep{kang2023enhancing,kang2023learning}, highlighting the generality of the approach.

The key advantage of DP over soft-constrained (PINN) and purely data-driven (NODE) paradigms lies in its enforcement of governing equations as hard constraints within the computational graph,
while retaining end-to-end differentiability. 
This reduces the reliance on loss balancing and promotes physically consistent gradient propagation through the solver--a property that our benchmark demonstrates to be important for control synthesis.

\subsection{Classical Approaches for System Identification}

For completeness, we briefly review classical (non-ML) methods for system identification in power systems. 
The extended Kalman filter and unscented Kalman filter have been widely used for online parameter estimation \citep{ghahremanidynamic2011},
 providing recursive state and parameter estimates under Gaussian noise assumptions. 
Moving Horizon Estimation offers an optimization-based alternative that can incorporate state constraints \citep{huangmoving2013}. 
Particle filtering methods have also been explored for nonlinear state estimation in power systems,
 particularly in the context of dynamic state estimation under non-Gaussian noise \citep{kozierski2015particle}.

These classical approaches serve as established baselines for system identification. 
However, they differ fundamentally from the differentiable paradigms considered in this work, as they do not enable end-to-end gradient-based learning via automatic differentiation. 
Our focus is on comparing differentiable frameworks that support both parameter estimation and downstream control synthesis within a unified computational pipeline.

\section{Problem Formulation}
\label{section:physical-model}
A power transmission grid consists of generators, loads, and transmission lines that interconnect them. Each generator is modeled as an internal node connected to a terminal bus through transient reactance, whereas load buses are connected only to the transmission network. Both generators and loads exhibit their own intrinsic dynamics that are coupled through nonlinear AC power-flows couplings. 
The standard structure-preserving model \citep{bergen2007structure} explicitly represents both generator and load dynamics while retaining the original network topology and nonlinear AC power-flow couplings. This model has been widely adopted as a reference for stability and control analysis in power systems \citep{vuframework2017,zoupinnsdriven2025}.

The dynamics of a single generator is described by the swing equation, which governs the evolution of the rotor (internal voltage) angle $\delta$: 
\begin{equation}
\label{eq:swing}
    M \DD{^2\delta}{t^2} + D \DD{\delta}{t} + P_{e}(\delta) = P_{m}.
\end{equation}
Here, $M$ denotes the inertia constant, 
$D$ is the damping coefficient, 
and $P_{e}(\delta)$ and $P_{m}$ represent the electrical output 
and mechanical input powers, respectively. 

\subsection{Single Machine Infinite Bus system}

\begin{figure}[htbp] 
    \centering
    
    \tikzset{every picture/.style={line width=0.75pt}}
    \begin{tikzpicture}[x=0.75pt,y=0.75pt,yscale=-1,xscale=1]
    
\draw   (323.33,99.67) -- (337.73,99.67) .. controls (337.94,89.16) and (339.94,80.18) .. (342.77,77.03) .. controls (345.6,73.89) and (348.68,77.23) .. (350.53,85.45) .. controls (351.96,91.86) and (352.55,100.15) .. (352.13,108.2) .. controls (352.13,111.33) and (351.42,113.88) .. (350.53,113.88) .. controls (349.65,113.88) and (348.93,111.33) .. (348.93,108.2) .. controls (348.52,100.15) and (349.1,91.86) .. (350.53,85.45) .. controls (352.2,78.62) and (354.51,74.75) .. (356.93,74.75) .. controls (359.36,74.75) and (361.67,78.62) .. (363.33,85.45) .. controls (364.76,91.86) and (365.35,100.15) .. (364.93,108.2) .. controls (364.93,111.33) and (364.22,113.88) .. (363.33,113.88) .. controls (362.45,113.88) and (361.73,111.33) .. (361.73,108.2) .. controls (361.32,100.15) and (361.9,91.86) .. (363.33,85.45) .. controls (365,78.62) and (367.31,74.75) .. (369.73,74.75) .. controls (372.16,74.75) and (374.47,78.62) .. (376.13,85.45) .. controls (377.56,91.86) and (378.15,100.15) .. (377.73,108.2) .. controls (377.73,111.33) and (377.02,113.88) .. (376.13,113.88) .. controls (375.25,113.88) and (374.53,111.33) .. (374.53,108.2) .. controls (374.12,100.15) and (374.7,91.86) .. (376.13,85.45) .. controls (377.99,77.23) and (381.07,73.89) .. (383.9,77.03) .. controls (386.73,80.18) and (388.73,89.16) .. (388.93,99.67) -- (403.33,99.67) ;
\draw   (172.33,99.67) .. controls (172.33,85.86) and (183.53,74.67) .. (197.33,74.67) .. controls (211.14,74.67) and (222.33,85.86) .. (222.33,99.67) .. controls (222.33,113.47) and (211.14,124.67) .. (197.33,124.67) .. controls (183.53,124.67) and (172.33,113.47) .. (172.33,99.67) -- cycle ;
\draw   (178.84,99.67) .. controls (186.36,83.3) and (189.91,83.22) .. (197.33,99.67) .. controls (204.76,116.12) and (208.24,116.22) .. (215.84,99.67) ;
\draw    (273,68.67) -- (273,132) ;
\draw    (403.33,99.67) -- (444,99.67) ;
\draw   (444,99.67) .. controls (444,85.86) and (455.19,74.67) .. (469,74.67) .. controls (482.81,74.67) and (494,85.86) .. (494,99.67) .. controls (494,113.47) and (482.81,124.67) .. (469,124.67) .. controls (455.19,124.67) and (444,113.47) .. (444,99.67) -- cycle ;
\draw    (222.33,99.67) -- (323.33,99.67) ;

\draw (454.67,89.4) node [anchor=north west][inner sep=0.75pt]  [font=\Large]  {$V_{\infty }$};
\draw (352.33,46.73) node [anchor=north west][inner sep=0.75pt]  [font=\Large]  {$X$};
\draw (240.67,70.07) node [anchor=north west][inner sep=0.75pt]  [font=\Large]  {$E$};

    \end{tikzpicture}

    \caption{Single machine infinite bus (SMIB) system.}
    \label{fig:smib}
\end{figure}

As illustrated in Fig.~\ref{fig:smib}, 
the SMIB system serves as a simplified representation of the structure-preserving model, 
connecting a single synchronous generator to an infinite bus that approximates the remainder of the power grid. The infinite bus provides a fixed reference for the generator by maintaining a constant voltage magnitude and frequency \citep{sauer2017power}. Assuming negligible transmission losses and bus voltage deviations, the electrical output power of the generator is given by $$P_{e}(\delta)=\frac{EV_\infty}{X} \sin \delta$$ 
where $E$ is the internal generator voltage, 
$V_\infty$ is the infinite bus voltage,
and $X$ denotes the equivalent reactance between the generator and the infinite bus. Substituting this expression into (\ref{eq:swing}) yields 
the swing equation for the SMIB system:
\begin{equation}
\label{eq:smib}
    \underbrace{M \DD{^2\delta}{t^2}}_{\text{inertia}} +
    \underbrace{D \DD{\delta}{t}}_{\text{damping}} + 
    \underbrace{\frac{EV_\infty}{X} \sin \delta}_{\text{restoring}} = P_{m}.
\end{equation}

Equation~(\ref{eq:smib}) admits a direct mechanical analogy to a mass--spring--damper system. Under this interpretation, the rotor angle $\delta$ corresponds to the displacement of a mass, 
the inertia constant $M$ represents the mass that resists changes in rotor speed, and 
the damping coefficient $D$ accounts for the dissipation of oscillatory energy. Moreover, the synchronizing stiffness, defined as $$\DD{P_e}{\delta}=\frac{EV_\infty}{X} \cos \delta,$$ acts as an effective spring constant that pulls the rotor angle back toward equilibrium.
 
By introducing the rotor speed $\omega$, we reformulate (\ref{eq:smib}) as an equivalent first-order state-space system: 
\begin{subequations}
\label{eq:smib2}
\begin{align}
    \DD{\delta}{t} &= \omega, \\
    \DD{\omega}{t} &= \frac{1}{M} \LRp{P_m - \frac{EV_\infty}{X}\sin \delta - D\omega}.
\end{align}
\end{subequations}

Defining the state vector $\xb= (\delta, \omega)^T$ and the control input $u=P_m$, we compactly rewrite (\ref{eq:smib2}) as 
\begin{equation}
\label{eq:smib-compact}
    \DD{\xb}{t} = \fb(\xb,u)
\end{equation}
where $\fb(\xb,u)=\frac{1}{M}\LRp{M\omega,P_m - \frac{EV_\infty}{X}\sin \delta - D\omega}^T$.

As power systems transition toward renewable-dominated grids, 
the passive nature of conventional grid-following inverters becomes insufficient for maintaining system stability. Low-inertia renewable energy system, including wind and solar power generation, are particularly susceptible to rapid dynamic fluctuations. Consequently, these systems must actively regulate their synchronization with the grid. 
To address this challenge, inverters must adopt grid-forming capabilities that emulate the behavior of conventional synchronous machines by synthesizing virtual inertia and damping responses \citep{prodanovichighquality2006,zhong2010synchronverters}. 

To fully leverage these capabilities, optimal control strategies such as LQR are often employed. However, 
the performance of LQR relies on an accurate knowledge of the systems' underlying physical parameters. Unlike traditional synchronous generators, where inertia constant ($M$) and damping coefficient ($D$) are fixed mechanical properties, the virtual inertia in modern inverter-dominated grids is often uncertain and time-varying
\citep{ratnam2020future}.

Consequently, the adoption of grid-forming control is insufficient; it must be complemented by the accurate identification of the dynamic parameters from real-time measurements. 
This requirements motivates the use of differentiable modeling techniques, which not only enable robust parameter estimation under noisy conditions, but also facilitate the synthesis of stable and optimal control gains. 

To formulate this identification problem, we treat $M$ and $D$ in (\ref{eq:smib2}) as learnable parameters $\theta_M$ and $\theta_D$, respectively: 
\begin{subequations}
\label{eq:smib-inverse}
\begin{align}
    \DD{\delta}{t} &= \omega, \\
    \DD{\omega}{t} &= \frac{1}{\theta_M} \LRp{P_m - \frac{EV_\infty}{X}\sin \delta - \theta_D\omega}.
\end{align}
\end{subequations}

\subsection{LQR Control}
  
LQR is a state-feedback control method that synthesizes optimal gains from a linearized system model, explicitly balancing stability and control effort. Its performance critically depends on the accuracy of the system Jacobians used in the linearization \citep{markoviclqrbased2019,lvlqr2024}.

For the SMIB system, the small-signal dynamics linearized around the equilibrium point $\xb^* = (\delta^*, \omega^*)^T$ are described by the state-space equation: 
\begin{equation}
\label{eq:smib-linearized}
    \DD{\Delta\xb}{t} = A\Delta\xb + B\Delta u
\end{equation}
where $\Delta\xb = (\Delta\delta, \Delta\omega)^T$ represents the state perturbations from the equilibrium, and $\Delta u = \Delta P_m$ denotes the control input deviation. The system Jacobian matrix $A$ and the control input matrix $B$ are given by  
\begin{align*}
A = 
\begin{pmatrix}
0 & 1 \\
-\frac{EV_\infty}{MX} \cos \delta^* & -\frac{D}{M}
\end{pmatrix}  
\quad \text{and} \quad 
B = 
\begin{pmatrix}
0  \\
\frac{1}{M}
\end{pmatrix}.
\end{align*}

The LQR objective is to minimize the quadratic cost function 
\begin{align}
    \mathcal{C}=\int_{0}^{\infty} \Delta \xb^T Q\Delta \xb + \Delta u^T R \Delta u ~dt,
\end{align}
where $Q$ and $R$ are symmetric positive semi-definite and positive weighting matrices, respectively. 
The optimal control input is given by $\Delta u=-K \Delta \xb$, where the feedback gain $K=R^{-1} B^T P$ is derived from the solution $P$ of the continuous-time algebraic Riccati equation:
$$A^T P + PA - PB R^{-1}B^T P + Q = 0.$$
In the closed-loop simulation, the control input $\Delta u$ is superimposed on the nominal mechanical power $P_m$ in (\ref{eq:smib2}). Specifically, the input is updated at each time step as $P_m\leftarrow P_m + \Delta u$.

\section{Differentiable Models}
\label{section:differentiable-model}

In this section, we detail the three differentiable modeling paradigms applied to the SMIB system described in (\ref{eq:smib-compact}). Each paradigm encodes physical knowledge at a different level of structural commitment, as introduced in Section~\ref{section:related-work}: Specifically, NODEs learn the dynamics operator from data with minimal structural assumptions, PINNs integrate physics as soft constraints in the loss function, and DP adopts a hybrid physics-neural strategy that enforces the governing equations as hard constraints within the computational graph while optimizing unknown parameters via backpropagation. Fig.~\ref{fig:paradigm_comparison} illustrates the differences in their knowledge encoding strategies.

\begin{figure}[t]
\centering
\setlength{\unitlength}{1cm}
\begin{picture}(8.5,2.0)
\put(0,1.5){\makebox(0,0)[l]{\footnotesize\textbf{Weak} knowledge}}
\put(7.0,1.5){\makebox(0,0)[l]{\footnotesize\textbf{Strong} knowledge}}
\put(0.0,1.0){\vector(1,0){8.4}}
\put(0.5,0.3){\framebox(1.8,0.5){\footnotesize NODE}}
\put(3.3,0.3){\framebox(1.8,0.5){\footnotesize PINN}}
\put(6.1,0.3){\framebox(1.8,0.5){\footnotesize DP}}
\put(0.5,0.0){\makebox(1.8,0.2){\tiny Data-driven}}
\put(3.3,0.0){\makebox(1.8,0.2){\tiny Soft-constrained}}
\put(6.1,0.0){\makebox(1.8,0.2){\tiny Hard-constrained}}
\end{picture}
\caption{The knowledge integration spectrum. The three paradigms encode physical knowledge at increasing levels of structural commitment, from purely data-driven (NODE) to soft-constrained (PINN) and hard-constrained (DP) formulations.}
\label{fig:paradigm_comparison}
\end{figure}

\subsection{Neural Ordinary Differential Equations}
From a knowledge integration perspective,
NODEs represent a \emph{data-driven} paradigm with a minimal structural prior: the assumption that system dynamics follow a continuous-time ODE. This inductive bias—--expressed as $\dot{\xb} = \fb(\xb, u)$--—captures the existence of an underlying dynamical law without imposing a specific functional form. 
NODEs are therefore particularly suitable for settings where the governing equations are unknown and sufficient data are available.
Unlike direct state approximation methods, NODEs capture the SMIB dynamics by learning the vector field (tendency function). 
Specifically, $\fb(\xb,u)$ in (\ref{eq:smib-compact}) is approximated  
by a neural network $\hat{\fb}(\xb,u; \theta)$, 
leading to the parameterized ordinary differential equation 
\begin{equation}
\label{eq:node}
    \DD{\xb}{t}=\hat{\fb}(\xb,u; \theta). 
\end{equation}
Given an initial condition, the state trajectory is obtained by numerically integrating (\ref{eq:node}) using a standard time integrator, such as a Runge-Kutta method. 
Training is performed by minimizing the discrepancies between simulated and observed trajectories. Specifically, we randomly select $n_b$ trajectory segments. For each sample $i$, we generate $m$ consecutive predictions $\hat{\xb}_1^{(i)}, \cdots, \hat{\xb}_m^{(i)}$ starting from an initial state $\xb_0^{(i)}$. 
The neural network parameters $\theta$ are optimized using the following loss function: 
\begin{equation}
\label{eq:loss-node}
    \mathcal{L} = \frac{1}{n_b m} \sum_{i=1}^{n_b} \sum_{j=1}^{m} \norm{\hat{\xb}_j^{(i)} -\xb_j^{(i)}}^2.
\end{equation}
This data-driven approach allows NODEs to identify system dynamics directly from data, providing a flexible framework for modeling complex systems without requiring explicit knowledge of the governing equations.
However, this flexibility comes at a cost: standard NODEs are inherently black-box models and lack the structural transparency required to explicitly recover physical parameters. 
This limitation motivates the shift toward DP, which mitigates these shortcomings.

\subsection{Physics-Informed Neural Networks}
 PINNs, in contrast, represent a \emph{soft-constrained} paradigm: they encode domain knowledge by penalizing violations of governing equations in the loss function, 
thereby encouraging---but not guaranteeing--- physical consistency.
PINNs directly approximate the state trajectories $\xb(t)$ of the SMIB system by neural network functions $\hat{\xb}(t; \theta)$, where $\theta$ denotes the network parameters.
To ensure physical consistency, PINNs enforce the underlying SMIB dynamics by embedding the differential equations (\ref{eq:smib-compact}) directly into the training objective. The network is optimized by minimizing the following physics-informed loss function, which penalizes the residuals of the system dynamics at a set of collocation points:
\begin{equation}
\label{eq:pinn-loss}
    \mathcal{L}_{phys} = \frac{1}{n_c} \sum_{i=1}^{n_c} \norm{\DD{\hat{\xb}_i}{t} - \fb(\hat{\xb}_i)}^2.
\end{equation}
Here, $\hat{\xb}_i=\hat{\xb}(t_i; \theta)$ denotes the predicted state at time $t=t_i$, and $n_c$ is the number of collocation points. 

In practice, the loss function in (\ref{eq:pinn-loss}) can be augmented with additional terms penalizing discrepancies between observed data and model predictions and enforcing the prescribed initial conditions. These constraints enable the PINN to simultaneously satisfy the governing equations (\ref{eq:smib-compact}) and fit the available measurement data:
\begin{equation}
\label{eq:pinn-loss2}
    \mathcal{L} = \mathcal{L}_{phys}  
    + \lambda_{d} \frac{1}{n_d} \sum_{i=1}^{n_d} \norm{\hat{\xb}_i -\xb_i}^2 
    + \lambda_i \norm{\hat{\xb}_0 -\xb_0}^2.
\end{equation}
Here, $n_d$ denotes the number of data points, while  $\lambda_d$ and $\lambda_i$ are the weighting factors for the data mismatch and the initial condition losses, respectively. 

For parameter identification, 
the unknown parameters $\theta_M$ and $\theta_D$ appearing in (\ref{eq:smib-inverse}) are treated as trainable variables and optimized simultaneously with the network parameters $\theta$ during training.

\subsection{Differentiable Programming: A Hybrid Physics-Neural Approach}

DP represents a \emph{hard-constrained, hybrid} paradigm that bridges traditional physics-based modeling and deep learning. Unlike NODEs or PINNs, DP retains the governing equations as the core computational backbone and leverages gradient-based optimization to identify unknown physical parameters. The result is a hybrid architecture in which the \emph{model structure} is derived from classical physics, while the \emph{learning algorithm} is driven by gradient-based training. In this framework, the only learnable quantities are the unknown physical parameters—--in our case, the inertia constant $M$ and the damping coefficient $D$. This maximal structural commitment reduces the need for loss balancing and promotes physically consistent gradient propagation.

Concretely, the DP approach treats the numerical simulation of the SMIB system in (\ref{eq:smib-compact}) as an end-to-end differentiable computational graph. The ODE solver itself serves as a differentiable layer through which gradients are backpropagated to update the unknown parameters. Forward predictions are generated by numerically integrating (\ref{eq:smib-inverse}) and evaluated using the loss function in (\ref{eq:loss-node}). The differentiability of the solver enables efficient and consistent parameter updates.

A direct consequence of this construction is that DP provides system Jacobians and sensitivities with respect to both states and parameters through automatic differentiation, avoiding the approximation errors inherent in PINNs (which approximate solution trajectories) and NODEs (which approximate the vector field). This property is particularly valuable for control-oriented analysis, where accurate sensitivities are required for gain synthesis and stability assessment.

\section{Numerical Results}
\label{section:numerical-results}

In this section, we evaluate the three knowledge integration paradigms---NODE (data-driven), PINN (soft-constrained), and DP (hard-constrained)---on three complementary tasks of increasing difficulty.
We begin by evaluating forward prediction accuracy to assess the generalization performance.
Next, we consider an inverse problem to identify unknown physical parameters ($M$ and $D$) from data, thereby testing the efficiency of each paradigm for knowledge extraction.
Finally, we examine control-oriented applicability through LQR synthesis, evaluating how knowledge fidelity propagates to downstream tasks. 
All models are implemented using PyTorch on an NVIDIA RTX 4500 Ada Generation GPU. The differential equation solvers use the \texttt{torchdiffeq} library, and the algebraic Riccati equations for LQR synthesis are solved with \texttt{scipy.linalg} library.

Both the PINN and NODE models employ fully connected neural networks with four hidden layers of sizes $[200,150,100,50]$ and $\tanh$ activations. The PINN model maps the time coordinate $t$ (1 neuron) to the state vector $\xb$ (2 neurons), whereas the NODE model maps the state $\xb$ (2 neurons) to the time derivative $\dot{\xb}$ (2 neurons). For parameter estimation, both the DP and PINN models treat the physical coefficients $\theta_M$ and $\theta_D$ as learnable scalar parameters. For the control task, the NODE input is augmented to include the control input $u$, yielding a network with input dimension 3 ($\xb$ and $u$) and output dimension 2 ($\dot{\xb}$).

To accommodate the structural differences among the models, we adopt distinct data sampling strategies. The PINN model is trained on $n_c=1000$ random collocation points, whereas the NODE and DP models use a fixed set of $n_b=100$ trajectory segments with sequence length $m$. All models are trained using the Adam optimizer with fixed learning rates. The differential equations are integrated using the \texttt{dopri5} solver \citep{hairer1993solving} with tolerances \texttt{rtol=1e-7} and \texttt{atol=1e-9}. Further details on hyperparameter configurations are provided in the Appendix.

We quantify the model accuracy using the relative $\ell_2$ error, defined as $\frac{\|\hat{\mathbf{v}}-\mathbf{v}_{r}\|_2}{\|\mathbf{v}_r\|_2}$, where $\hat{\bf v}$ and ${\bf v}_r$ denote the model prediction and the reference solution, respectively.

\subsection{Forward problem: Trajectory Prediction}

In this section, we evaluate the capability of NODE and PINN to capture the underlying dynamics of the SMIB system under two regimes: stable and oscillatory. 
We assume that the system is initialized away from equilibrium.
The physical parameters for the stable regime are configured as $P_m=0.1$, $E=V_\infty=1$, $X=5$, $D=0.2$, and $M=0.4$, while the oscillatory regime uses $P_m=0.5$, $E=X=5$, $V_\infty=1$, $D=0.01$, and $M=0.1$. Using these physical parameters and the initial condition $\xb =(0.1,0.1)^T$, we integrate (\ref{eq:smib2}) using the \texttt{dopri5} method from $t=0$ to $t=20$ with a timestep size of $dt=0.02$ to obtain the reference solutions.

\begin{figure}[!t]
\centering
\includegraphics[trim=0cm 0cm 0cm 0cm,clip=true,width=0.95\linewidth]{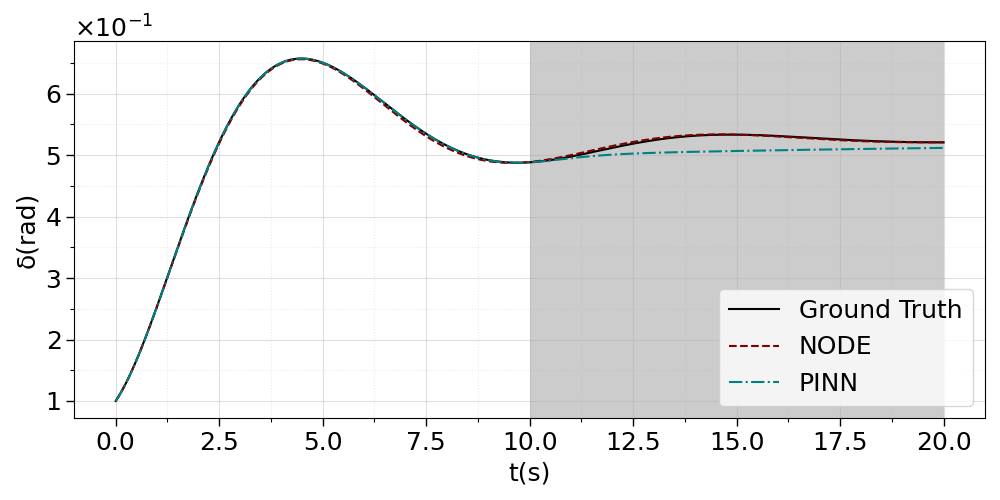}%
\\
\includegraphics[trim=0cm 0cm 0cm 0cm,clip=true,width=0.95\linewidth]{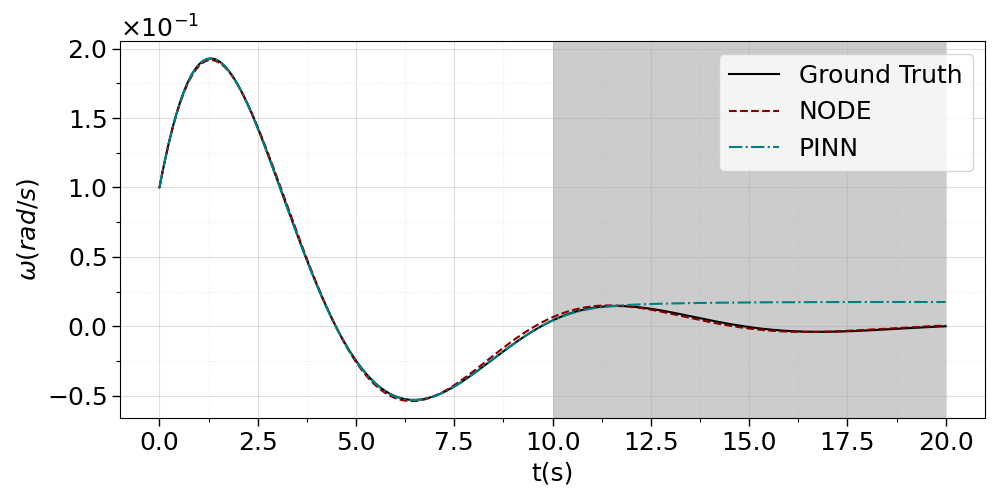}%
\caption{Comparison of predicted trajectories for rotor angle $\delta$ (top) and angular velocity $\omega$ (bottom) in the stable regime. The NODE model accurately captures the underlying dynamics and closely follows the ground truth, whereas the PINN model fails to generalize in the extrapolation regime. The shaded region ($t>10$~s) indicates the extrapolation horizon.}
\label{fig:smooth-trajectory}
\end{figure}

\begin{figure}[!t]
\centering
\includegraphics[trim=0cm 0cm 0cm 0cm,clip=true,width=0.95\linewidth]{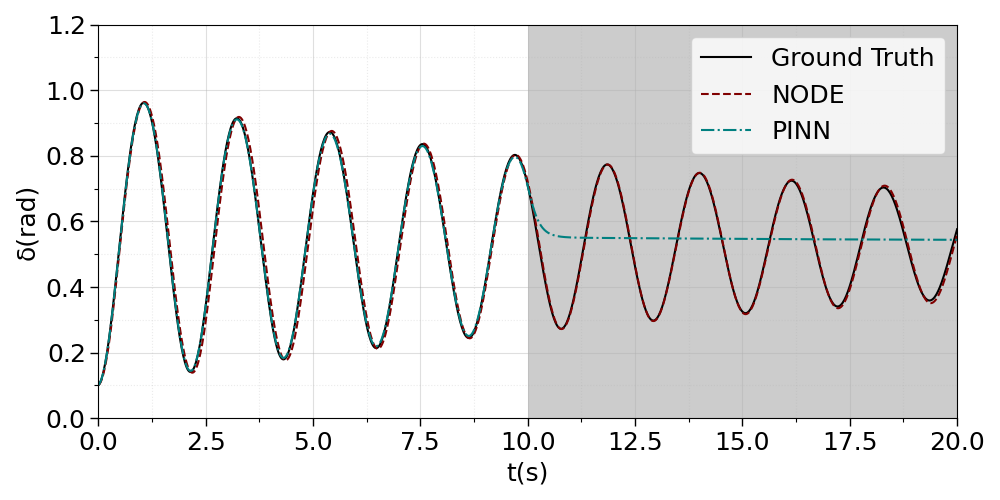}%
\\
\includegraphics[trim=0cm 0cm 0cm 0cm,clip=true,width=0.95\linewidth]{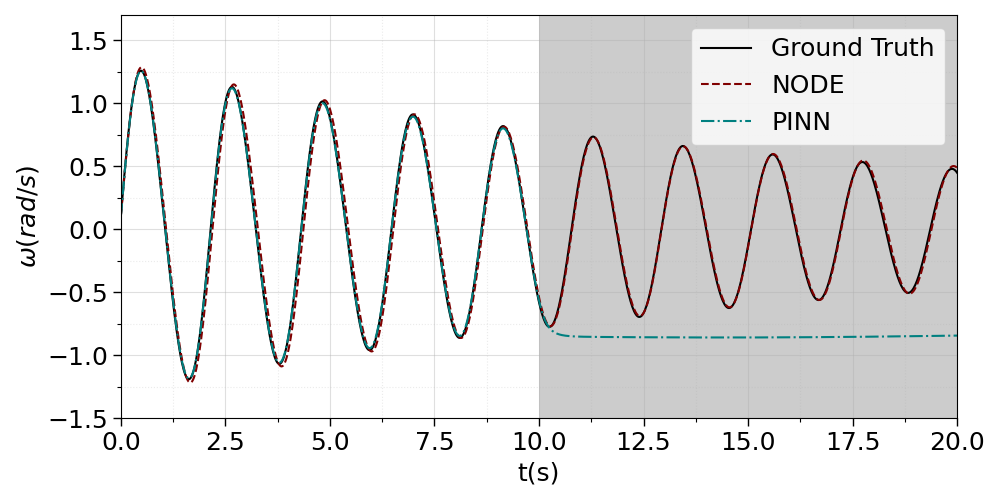}%
\caption{Comparison of predicted trajectories for rotor angle $\delta$ (top) and angular velocity $\omega$ (bottom) in the oscillatory regime.
Compared with the stable regime (Fig.~\ref{fig:smooth-trajectory}), 
extrapolation is markedly more challenging: PINN model fails to generalize soon after $t=10~s$, while NODE continues to track the underlying dynamics. The shaded region ($t>10$~s) indicates the extrapolation horizon.}
\label{fig:oscillation-trajectory}
\end{figure}

Fig.~\ref{fig:smooth-trajectory} and Fig.~\ref{fig:oscillation-trajectory} compare the predicted state trajectories obtained using the NODE and PINN models. Fig.~\ref{fig:smooth-trajectory} corresponds to the stable regime, while Fig.~\ref{fig:oscillation-trajectory} illustrates the oscillatory regime. 
The models are trained on the interval $t\in[0,10]$, and 
perform predictions over the full time horizon $t\in[0,20]$, where the shaded region indicates the extrapolation region. 
 While all models perform well within the training region, PINN struggles to predict the state variables $\delta$ and $\omega$ accurately during extrapolation. In contrast, NODE exhibits superior generalization capabilities. This performance gap stems from a fundamental difference in methodology: PINN approximates the solution map $\hat{\xb} (t)$, whereas NODE explicitly learns the underlying continuous-time vector field $\hat{\fb}(\xb)$. By learning the intrinsic dynamics rather than fitting a trajectory, NODE provides more robust extrapolation behavior.

To quantify model accuracy, we perform numerical experiments over 10 independent trials with distinct random seeds.
Table~\ref{tab:l2err_forward} summarizes the mean and standard deviation of the relative errors for the state variables $\delta$ and $\omega$ across both stable and oscillatory regimes. Overall, NODE consistently outperforms PINN in all settings. In the stable regime, NODE reduces the average error by approximately a factor of five for both $\delta$ and $\omega$. In the oscillatory regime, the performance gap becomes even more pronounced: NODE achieves substantially lower errors, particularly for $\omega$, where the reduction exceeds an order of magnitude. These results indicate that NODE provides more accurate and robust predictions, especially in dynamically challenging regimes. 

\begin{table*}[t]
\centering
\caption{Comparison of relative $\ell_2$ errors in predicting $\delta$ and $\omega$ for PINN and NODE models}
\label{tab:l2err_forward}
\begin{tabular}{llcccccccc}
\toprule
\multirow{2}{*}{Regime} & \multirow{2}{*}{Model}
& \multicolumn{4}{c}{$\delta$}
& \multicolumn{4}{c}{$\omega$} \\
\cmidrule(lr){3-6}\cmidrule(lr){7-10}
& & Min & Average & Max & Std
  & Min & Average & Max & Std\\
\midrule
\multirow{2}{*}{Stable} & NODE & 2.823e-03 & 4.614e-03 & 8.732e-03 & 1.647e-03 & 1.250e-02 & 1.957e-02 & 3.408e-02 & 6.486e-03 \\
 & PINN & 1.323e-02 & 2.134e-02 & 3.357e-02 & 6.949e-03 & 9.475e-02 & 1.303e-01 & 1.544e-01 & 1.963e-02 \\
 \midrule
 \multirow{2}{*}{Oscillatory} & NODE & 3.014e-02 & 4.295e-02 & 7.056e-02 & 1.302e-02 & 7.876e-02 & 1.187e-01 & 2.000e-01 & 3.957e-02 \\
 & PINN & 1.862e-01 & 2.669e-01 & 4.403e-01 & 8.248e-02 & 8.995e-01 & 1.203e+00 & 1.498e+00 & 1.467e-01 \\
\bottomrule
\end{tabular}
\end{table*}

Next, we examine the ability of NODE to capture the underlying dynamics in the presence of measurement noise. We introduce Gaussian noise to both the stable and oscillatory trajectories as $\xb_{\text{noisy}}(t) = \xb(t)(1+\epsilon)$, where $\epsilon\sim\mathcal{N}(0,\sigma^2)$ and $\sigma$ denotes the noise level. 
In the following experiments, noise levels of \(\sigma\in \{0.01,0.05\}\) (i.e., $1\%$ and $5\%$) are considered. All numerical experiments are performed over 10 independent trials with distinct random seeds.
Table~\ref{tab:l2err_forward_noise} reports the mean and standard deviation of the relative $\ell_2$ errors in $\delta$ and $\omega$ across both the stable and oscillatory regimes.
As the noise level increases, the relative $\ell_2$ errors increase accordingly.
Moreover, the oscillatory regime consistently exhibits higher errors than the stable regime, with $\delta$ errors increasing by a factor of approximately 3--7 and $\omega$ errors showing a more moderate increase depending on the noise level.

\begin{table*}[t]
\centering
\caption{Relative $\ell_2$ errors in $\delta$ and $\omega$ predictions for the NODE model}
\label{tab:l2err_forward_noise}
\begin{tabular}{llcccccccc}
\toprule
\multirow{2}{*}{Regime} & \multirow{2}{*}{Noise Level}
& \multicolumn{4}{c}{$\delta$}
& \multicolumn{4}{c}{$\omega$} \\
\cmidrule(lr){3-6}\cmidrule(lr){7-10}
& & Min & Average & Max & Std
  & Min & Average & Max & Std\\
\midrule
\multirow{2}{*}{Stable} & $\sigma=0.01$ & 4.615e-03 & 7.936e-03 & 1.030e-02 & 1.828e-03 & 1.976e-02 & 3.308e-02 & 4.922e-02 & 8.723e-03 \\
 & $\sigma=0.05$ & 2.220e-02 & 3.408e-02 & 5.943e-02 & 1.027e-02 & 1.093e-01 & 1.854e-01 & 2.929e-01 & 6.406e-02 \\
\midrule
\multirow{2}{*}{Oscillatory} & $\sigma=0.01$ & 2.657e-02 & 5.810e-02 & 9.544e-02 & 1.864e-02 & 6.978e-02 & 1.563e-01 & 2.594e-01 & 5.309e-02 \\
 & $\sigma=0.05$ & 8.675e-02 & 1.093e-01 & 1.557e-01 & 2.073e-02 & 2.040e-01 & 2.905e-01 & 4.402e-01 & 6.920e-02 \\
\bottomrule
\end{tabular}
\end{table*} 

\begin{figure}[!t]
\centering
\includegraphics[trim=0cm 0cm 0cm 0cm,clip=true,width=0.95\linewidth]{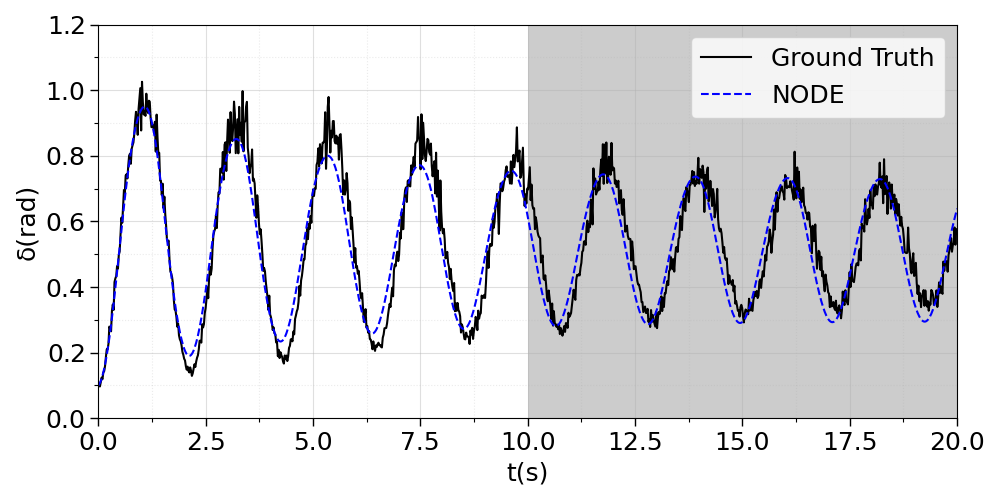}\\
\includegraphics[trim=0cm 0cm 0cm 0cm,clip=true,width=0.95\linewidth]{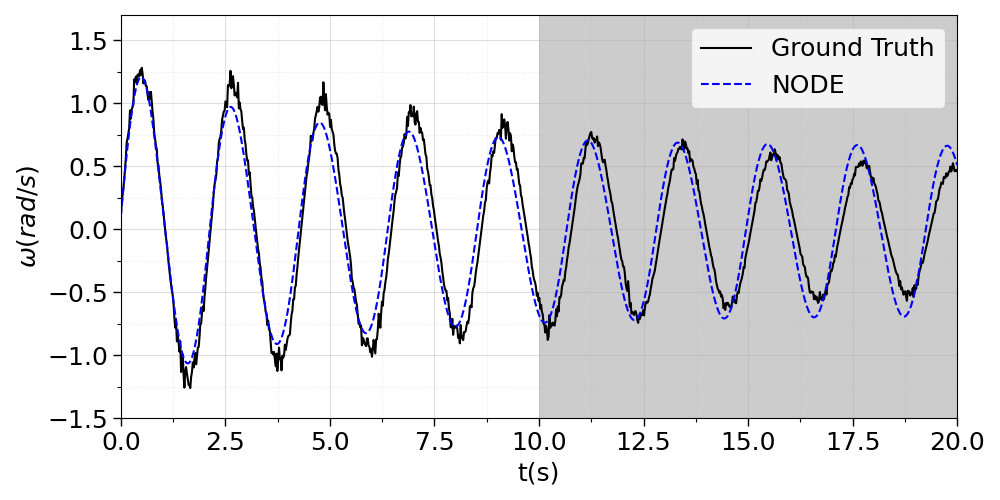}%
\caption{Predicted trajectories of NODE for rotor angle $\delta$ (top) and angular velocity $\omega$ (bottom) in the oscillatory regime with $5\%$ noise. The NODE prediction (blue dashed line) closely follows the ground truth (black solid line). The shaded region ($t > 10$) denotes the extrapolation regime beyond the training interval, demonstrating that NODE generalizes well under noisy observations.}
\label{fig:oscillation-trajectory-noise}
\end{figure}

Fig.~\ref{fig:oscillation-trajectory-noise} illustrates the predicted trajectories of NODE for $\delta$ and $\omega$ in the oscillatory regime under $5\%$ noise. Notably, NODE accurately recovers the underlying continuous dynamics directly from noisy observations, without relying on any explicit filtering. 

\subsection{Inverse problem: Discovery of Inertia and Damping}

We evaluate the ability of PINN and DP models to identify the inertia constant $M$ and damping coefficient $D$ from observed state trajectories. 
We consider the oscillatory regime, where the true values are $(M,D)=(0.1,0.01)$.
To ensure positivity ($M,D>0$), we parameterize both as softplus function $\ln(1+e^\theta)$, where $\theta$ denotes the unconstrained trainable parameter. 

\begin{figure}[!t]
\centering
\includegraphics[trim=0cm 0cm 0cm 0cm,clip=true,width=0.9\linewidth]{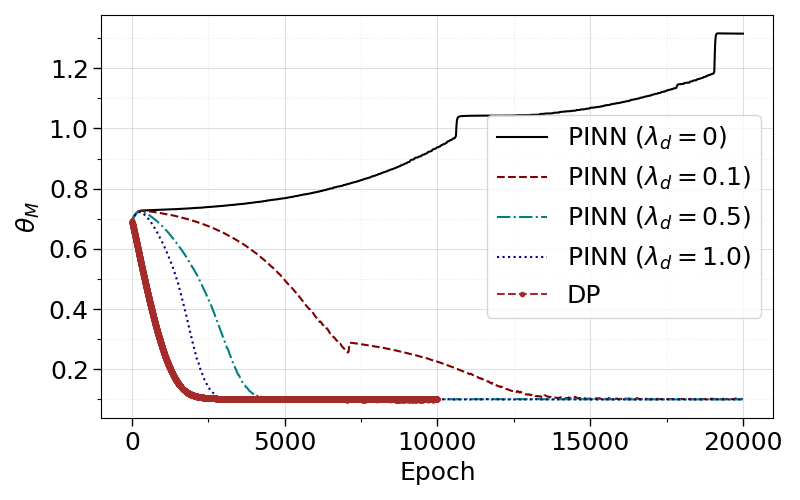}%
\\
\includegraphics[trim=0cm 0cm 0cm 0cm,clip=true,width=0.9\linewidth]{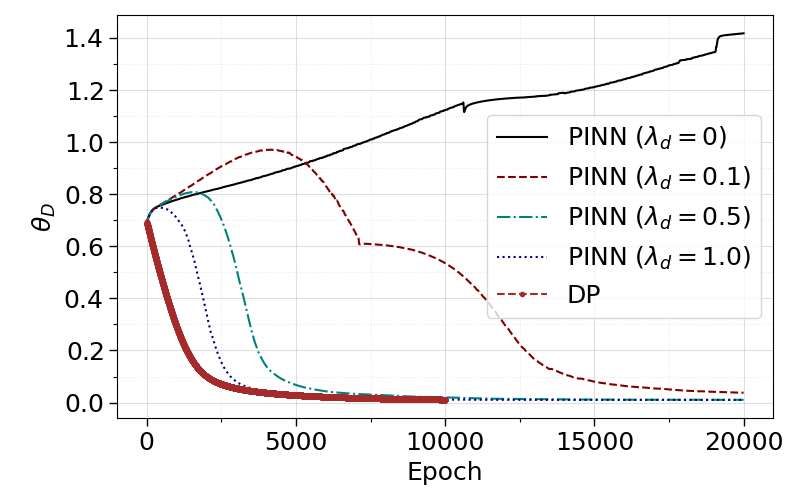}%
\caption{Convergence behavior of parameter estimation for inertia $\theta_M$ (top) and damping coefficient $\theta_D$ (bottom) with respect to the data loss weight $\lambda_d$. 
Increasing $\lambda_d$ accelerates the convergence of PINN toward the true values. The DP model converges substantially faster, requiring considerably fewer epochs to reach accurate estimates. } 
\label{fig:convergence_behavior}
\end{figure}

We first examine
the convergence behavior of the PINN and DP models in the noiseless setting ($\sigma=0$).  
Under a fixed learning rate of $10^{-3}$, Fig.~\ref{fig:convergence_behavior} shows that increasing the data loss weight $\lambda_d$ in PINN accelerates convergence toward the true values.
Notably, DP achieves faster convergence compared to PINN. This is because DP optimizes only two trainable parameters ($\theta_M$ and $\theta_D$), whereas PINN must simultaneously optimize a high-dimensional parameter space, including the neural network weights for $\delta$ and $\omega$.

\begin{table}[t]
\centering
\caption{Comparison of relative $\ell_2$ errors in estimating $\theta_M$ and $\theta_D$ for PINN and DP models under different noise levels}
\label{tab:l2err_inverse}
\begin{tabular}{llcc}
\toprule
Model & Noise Level & $\theta_M$ & $\theta_D$ \\
\midrule
\multirow{3}{*}{DP} & $\sigma=0$ & 1.610e-05 & 8.768e-03 \\
 & $\sigma=0.01$ & 7.307e-04 & 4.864e-02 \\
 & $\sigma=0.05$ & 1.593e-03 & 2.306e-01 \\
\midrule
\multirow{3}{*}{PINN} & $\sigma=0$ & 1.710e-03 & 1.321e-02 \\
 & $\sigma=0.01$ & 2.393e-03 & 1.280e-02 \\
 & $\sigma=0.05$ & 3.274e-03 & 2.414e-02 \\
\bottomrule
\end{tabular}
\end{table}

In real-world power systems, sensor measurements are often corrupted by noise. To account for this, we evaluate the robustness of the parameter estimation by introducing Gaussian noise into the oscillatory trajectories.
Table~\ref{tab:l2err_inverse} reports the 
relative $\ell_2$ errors in the estimated inertia constant $\theta_M$ and damping coefficient $\theta_D$, obtained from 5 independent trials. In this experiment, the data loss weight for PINN is set to $\lambda_d=1$. Overall, both models successfully recover the system parameters $M$ and $D$. As expected, the estimation error increases with the noise level. 
In the noiseless case ($\sigma=0$), DP achieves extremely high accuracy for $\theta_M$, with an error on the order of $10^{-5}$, compared to $10^{-3}$ for PINN, representing an improvement of approximately two orders of magnitude. For $\theta_D$, DP also outperforms PINN, although the gap is less pronounced.
As the noise level increases, both models exhibit a degradation in estimation accuracy. However, their behaviors differ significantly across parameters. For $\theta_M$, DP consistently maintains lower errors than PINN across all noise levels, indicating strong robustness in estimating the inertia parameter.
In contrast, for $\theta_D$, DP becomes increasingly sensitive to noise. At $\sigma=0.05$, the error for DP rises to $2.3\times10^{-1}$, which is an order of magnitude higher than that of PINN. Meanwhile, PINN demonstrates relatively stable performance for $\theta_D$, with only moderate increases in error as noise increases.
These results reveal a clear asymmetry in parameter identifiability: while DP excels at recovering $\theta_M$ with high accuracy, its estimation of $\theta_D$ is more susceptible to noise. This difference can be attributed to the weaker influence of damping on the system dynamics compared to inertia, making it more difficult to identify from noisy observations. In contrast, PINN exhibits more stable nominal values for $\theta_D$. We conjecture that this stability reflects partial compensation by the neural network, which can fit noisy trajectories with mildly inaccurate parameters by distorting its internal representation. A definitive analysis of this effect is beyond the scope of the present study.

\subsection{Control application: LQR Performance Comparison}

The efficacy of the control system is evaluated by its ability to stabilize the system under disturbance. We introduce a mechanical power perturbation $P_m^{\text{eff}} = P_m + \Delta P_m$ and analyze the resulting closed-loop response.
LQR control requires explicit knowledge of the system matrices $A$ and $B$ defined in (\ref{eq:smib-linearized}). However, since PINN learns a trajectory map rather than the underlying vector field, it does not directly provide these matrices. Furthermore, evaluating control responses under varying control inputs would require a control-parameterized solution map $\hat{\xb}(t,u;\theta)$, which significantly increases problem complexity. For this reason, PINNs are not considered in the control study.

In contrast, DP naturally supports control synthesis.
Once $\theta_M$ and $\theta_D$ are identified, 
the model can directly simulate closed-loop dynamics using embedded governing equations. 
To test the DP approach in a realistic identification setting,
we take the DP parameters estimated from the most challenging case in 
Table \ref{tab:l2err_inverse}---the noisy oscillatory regime at $\sigma=0.05$,
which yielded $\theta_M=0.1003$ and $\theta_D=0.0124$---and use them to construct the LQR control matrices $A$ and $B$. We then compare the resulting closed-loop behavior against that of an LQR controller designed using true parameters $M$ and $D$. This comparison isolates how much the noise-induced parameter error propagates into control performance. 
The equilibrium state is computed by setting the time derivatives in (\ref{eq:smib2}) to zero,
 yielding  $(\delta^*,\omega^*)=(0.5236, 0)$.
The weighting matrices are chosen as $Q = \mathrm{diag}(10, 1)$ and $R = 0.1$, assigning a higher penalty to rotor angle deviations compared to speed deviations, while allowing relatively aggressive control inputs. This choice emphasizes stability of the rotor angle, which is critical for maintaining synchronism, and facilitates a clear comparison of control performance across models.

\begin{figure}[!t]
\centering
\includegraphics[trim=0cm 0cm 0cm 0cm,clip=true,width=0.9\linewidth]{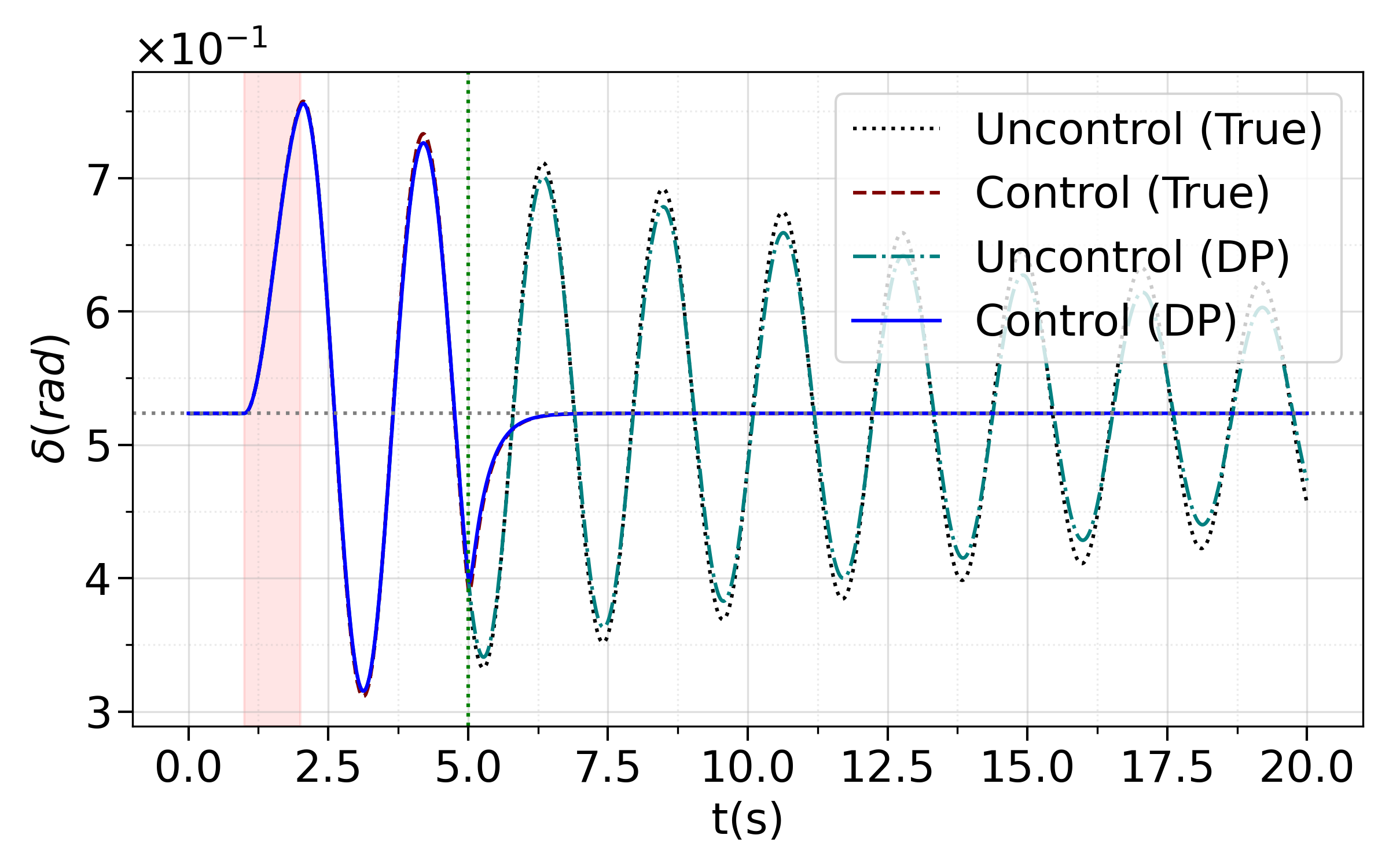}%
\\
\includegraphics[trim=0cm 0cm 0cm 0cm,clip=true,width=0.9\linewidth]{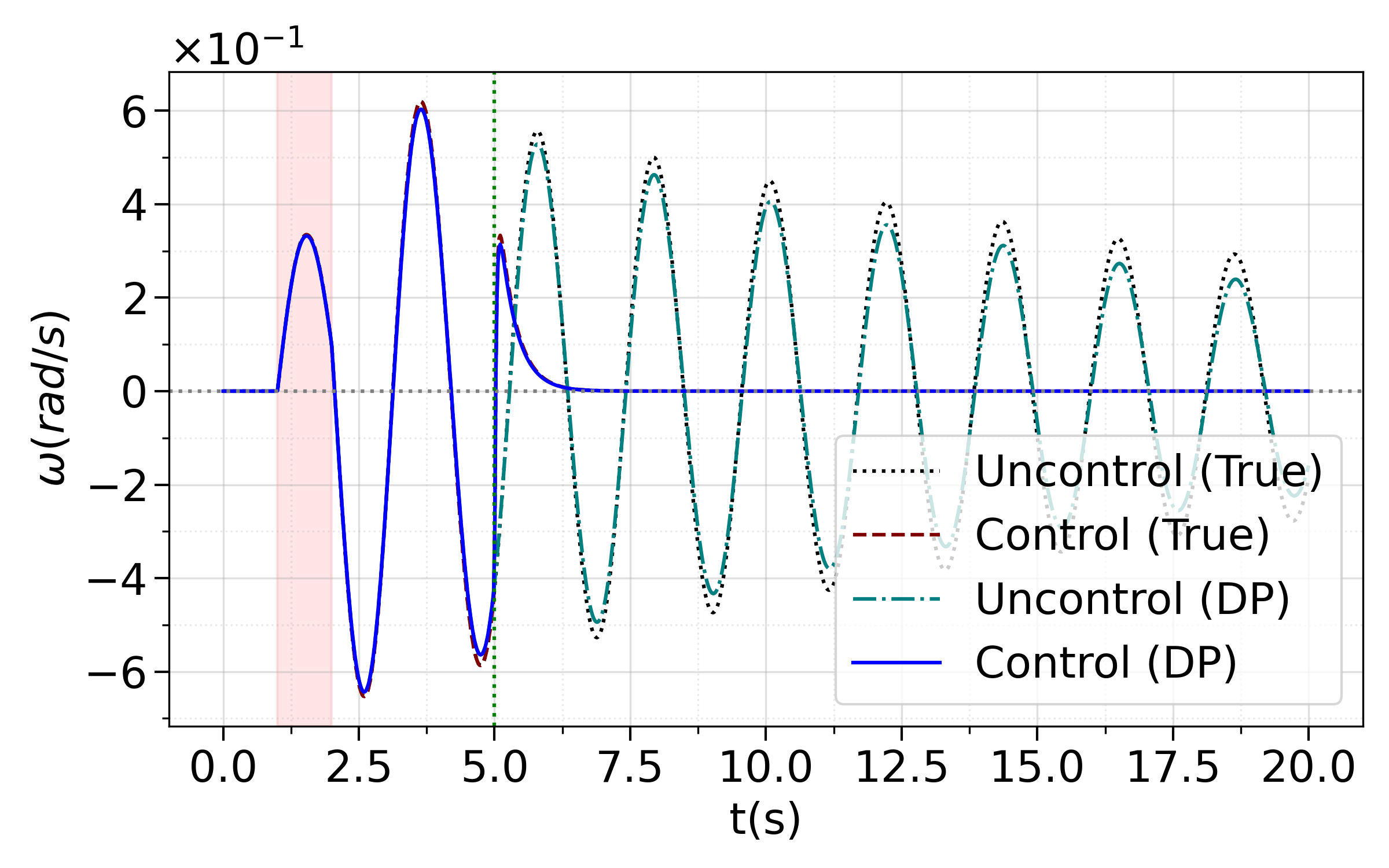}%
\caption{Comparison of LQR control performance for rotor angle $\delta$ (top) and angular velocity $\omega$ (bottom).
The trajectories compare the controller using exact system parameters $M$ and $D$ (Control (True)) against one using parameters estimated by the DP model, $\theta_M$ and $\theta_D$ (Control (DP)).
}
\label{fig:lqr-dp}
\end{figure}

Starting from the equilibrium initial condition $\xb=(0.5236,0)^T$, we simulate both the true system (\ref{eq:smib2}) and the DP model (\ref{eq:smib-inverse}) over the time interval $t\in[0,20]$. The physical parameters are set to $P_m=0.5$, $E=5$, $V_\infty=1$, $X=5$, $D=0.01$, and $M=0.1$. 
Fig.~\ref{fig:lqr-dp} compares the LQR control performance for rotor angle $\delta$ (top) and angular velocity $\omega$ (bottom). 
The trajectories obtained using the true parameters (Control (True)) are compared with those using the estimated parameters (Control (DP)). 
A mechanical power perturbation of $\Delta P_m = 0.1$ is applied over the interval $t\in [1,2]$ (pink shaded region), and the LQR controller is activated at $t=5$, indicated by the green dotted vertical line. 
The DP-based controller effectively stabilizes the system, producing trajectories that closely track those of the LQR controller designed with the true physical parameters.
Quantitatively, the relative $\ell_2$ errors are $0.0032$ for $\delta$ and $0.028$ for $\omega$. 
The larger relative error in $\omega$ is consistent with the identification results in Table~\ref{tab:l2err_inverse}: $\theta_D$ carries substantial error, and this parameter uncertainty propagates into the closed-loop response.

Importantly, the DP model preserves the same control-relevant structure as the governing equations. As a result, when the estimated parameters and corresponding Jacobians are sufficiently accurate, the resulting closed-loop trajectories closely match those obtained from the true-parameter controller, as quantified by the small trajectory errors above. These findings demonstrate that the DP approach bridges system identification and control synthesis by enforcing hard physical constraints, as defined in (\ref{eq:smib-inverse}).

We next address the control problem using the NODE model. 
Although the governing equations are assumed to be unknown in this setting, the matrices $A$ and $B$ in (\ref{eq:smib-linearized}) can be approximated from the learned vector field
$\hat{\fb}(\xb,u; \theta)$ as
$$
A\approx \dd{\hat{\fb}(\xb,u; \theta)}{\xb}, 
\quad \text{and} \quad 
B\approx \dd{\hat{\fb}(\xb,u; \theta)}{u}. 
$$
To capture the dependence of the dynamics on the control input $u$, we generate a training dataset consisting of four trajectories corresponding to $P_m=\{0.1, 0.2, 0.3, 0.4\}$. The physical parameters are set to $E=V_\infty=1$, $X=2$, $D=0.1$, and $M=0.2$. Using these parameters, we integrate (\ref{eq:smib2}) with the \texttt{dopri5} solver over the interval $t\in[0,20]$ with a timestep size of $dt=0.02$. The model is trained on the interval $t\in[0,10]$ using $n_b=100$ batches of sequence length $m=10$ per epoch, randomly sampled from the generated trajectories. 

\begin{figure}[!t]
\centering
\includegraphics[trim=0cm 0cm 0cm 0cm,clip=true,width=0.9\linewidth]{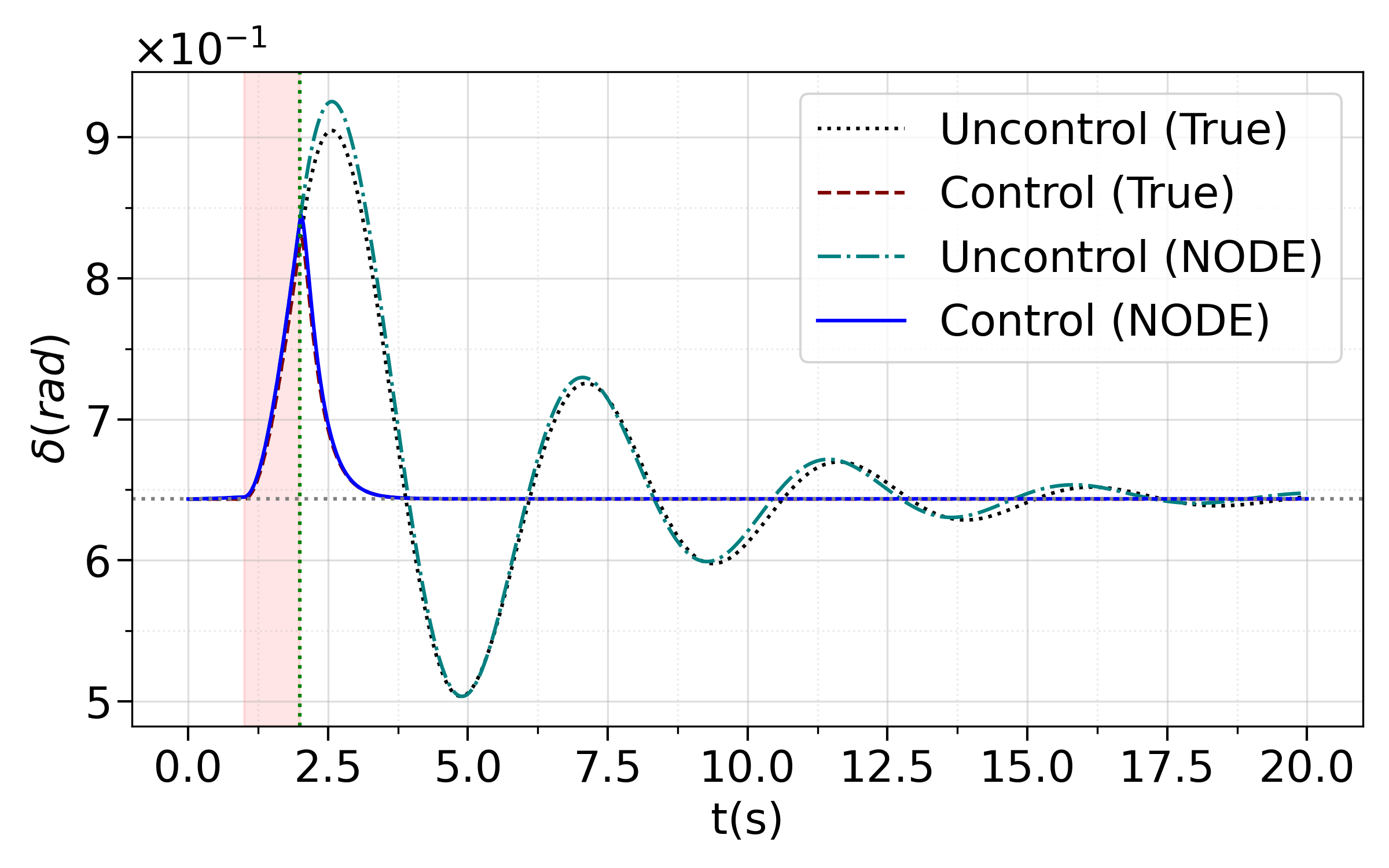}%
\\
\includegraphics[trim=0cm 0cm 0cm 0cm,clip=true,width=0.9\linewidth]{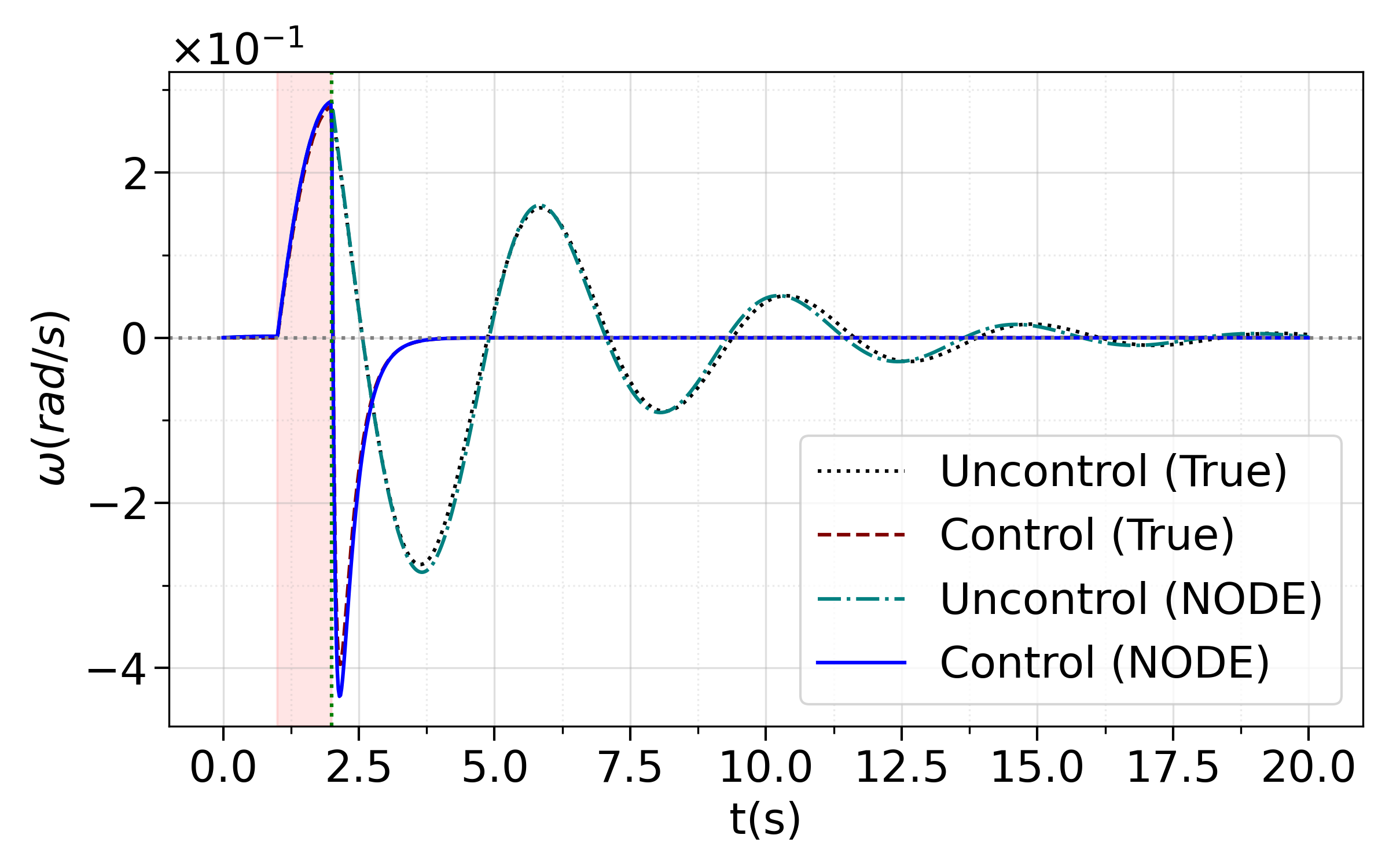}%
\caption{Comparison of LQR control performance for rotor angle $\delta$ (top) and angular velocity $\omega$ (bottom).
The trajectories compare the controller using the exact Jacobian (Control (True)) against one using the approximate Jacobian obtained from the NODE model (Control (NODE)).
}
\label{fig:lqr-node}
\end{figure}

Fig.~\ref{fig:lqr-node} shows the LQR control comparison for rotor angle $\delta$ (top) and angular velocity $\omega$ (bottom). 
The system is initialized at the equilibrium state $(\delta^*,\omega^*)=(0.6435, 0)$, corresponding to $P_m=0.3$. A perturbation ($\Delta P_m = 0.1$) is applied over the interval $t\in [1,2]$ (pink shaded region). The LQR controller, activated at $t=2$, rapidly suppresses the deviations in $\delta$ and $\omega$, restoring the system to equilibrium by approximately $t=3.7$.
These results indicate that NODE can support LQR synthesis even when the governing equations are unavailable. The control matrices $A$ and $B$ are obtained as derivatives of the learned vector field; the fidelity of the learned sensitivities therefore plays a central role in determining control performance, which we quantify next.

\begin{table}[!t]
\centering
\caption{Comparison of the true linearized SMIB system matrices against those recovered by NODE, evaluated at the equilibrium $(\delta^*, \omega^*) = (0.6435, 0)$ with $P_m = 0.3$. The LQR gain is computed with $Q = \mathrm{diag}(10, 1)$ and $R = 0.1$. Relative errors are computed using the Frobenius norm}
\label{tab:node-jacobian-accuracy}
\begin{tabular}{lccc}
\toprule
Matrix & True & NODE & Relative error \\
\midrule
$A$ 
& $\begin{pmatrix} 0 & 1 \\ -2 & -0.5 \end{pmatrix}$ 
& $\begin{pmatrix} -0.0071 & 1.0320 \\ -2.0651 & -0.5164 \end{pmatrix}$ 
& $3.26\%$ \\[1.2em]
$B$ 
& $\begin{pmatrix} 0 \\ 5 \end{pmatrix}$ 
& $\begin{pmatrix} 0.0186 \\ 5.1950 \end{pmatrix}$ 
& $3.92\%$ \\[1.2em]
$K$ (LQR gain)
& $\begin{pmatrix} 9.608 & 3.622 \end{pmatrix}$
& $\begin{pmatrix} 9.611 & 3.585 \end{pmatrix}$
& $0.36\%$ \\[0.3em]
\bottomrule
\end{tabular}
\end{table}

To quantify the fidelity of the learned sensitivities, we compare the NODE-derived control matrices against the analytical Jacobians of the linearized SMIB system in Table~\ref{tab:node-jacobian-accuracy}.
The NODE model recovers $A$ and $B$ with relative errors of $3.26\%$ and $3.92\%$, respectively.
Notably, these modeling errors propagate into a smaller error in the synthesized LQR gain: 
the true gain $K_{\text{true}}=(9.608,\, 3.622)$  and the NODE-based gain $K_{\text{NODE}}=(9.611,\, 3.585)$ differ by only $0.36\%$ in relative Frobenius norm. We report this as an empirical observation without claiming it as a generic property of Riccati-based synthesis. A systematic investigation of when and why such favorable propagation occurs---across different operating points, system dimensions, and learned-error structures---is left for future work.

At the same time, the learned matrices reveal the inductive-bias gap between NODE and DP. 
The structural entries of $A$ and $B$ are learned approximately rather than exactly by NODE. 
For example, $A_{11}$, which is identically zero in the true model, is recovered as $-0.0071$; 
$A_{12}$, which should be exactly one, is recovered as $1.0320$; 
and $B_1$, which should vanish, is recovered as $0.0186$. 
The DP model enforces these structural zeros and ones by construction, 
whereas NODE must learn them from data, at a small accuracy cost. 
This provides a concrete empirical illustration of the soft-versus-hard inductive-bias distinction discussed earlier. DP inherits exact structural relations from the governing equations, while NODE recovers them approximately without requiring any prior knowledge of the model form.
These observations underscore that the DP and NODE experiments are designed to illustrate two complementary regimes, not to benchmark the two approaches against each other. DP presumes that the governing equations are known and reduces control-oriented modeling to the identification of physical parameters; the resulting Jacobians are structurally exact, and the LQR gains inherit the guarantees of the linearized SMIB model. NODE, in contrast, assumes no such structural knowledge and must recover the control-relevant sensitivities $A$ and $B$ purely from trajectory data sampled across varying $P_m$. Demonstrating that both frameworks yield stabilizing LQR controllers---each within its own information regime---establishes that the choice between DP and NODE is governed not by control feasibility, but by the level of prior knowledge available about the system. In this sense, the two approaches are complementary: DP is the natural choice when a trusted physical model exists, while NODE provides a viable path to control synthesis when only observational data are available.

\section{Conclusion}
\label{section:conclusion}

This paper presented a systematic comparative study of three knowledge integration paradigms---data-driven (NODE), soft-constrained (PINN), and hard-constrained (DP)---for neural modeling of dynamical systems, evaluated through trajectory prediction, parameter identification, and control synthesis on a power system benchmark.

Our study yields three principal findings. First, the choice of knowledge representation fundamentally determines generalization: learning the system operator (NODE) enables robust extrapolation, whereas learning a solution map (PINN) restricts generalization to the training horizon. Second, the degree of structural commitment directly impacts inverse problem efficiency: hard-constrained formulations (DP) reduce the optimization to a low-dimensional parameter space, achieving substantially faster and more reliable convergence than soft-constrained approaches. Third, knowledge fidelity propagates to downstream tasks: the DP framework achieves control performance that closely matches the LQR controller designed using the true physical parameters, while NODE provides a viable data-driven alternative when governing equations are unavailable. In our SMIB example, NODE recovers control-relevant Jacobians with 3--4$\%$ relative error and yields LQR gains within $0.36\%$ of the ground truth. We report the latter as an empirical observation specific to our SMIB example and identify its systematic characterization as future work.

These findings are further reflected in the interpretability characteristics of each paradigm. DP preserves full structural transparency: the identified parameters retain direct physical meaning, and the model structure exactly mirrors the governing equations. Although NODE is not structurally interpretable in the same way as DP, it can be interrogated through Jacobian analysis to extract local linearizations for control design. In contrast, PINN offers limited interpretability, as the learned solution map does not explicitly expose the underlying dynamics operator.

Based on these observations, we summarize the trade-offs as a practical decision framework in Table~\ref{tab:decision_framework}. The framework intentionally focuses on the two extremes of fully known and fully unknown governing equations; the intermediate case of partial knowledge is discussed as a future direction below.

\begin{table}[t]
\centering
\caption{Decision framework for selecting knowledge integration paradigms based on available domain knowledge and task requirements}
\label{tab:decision_framework}
\renewcommand{\arraystretch}{1.25}
\setlength{\tabcolsep}{3pt}
\begin{tabular}{p{0.30\columnwidth}|p{0.60\columnwidth}}
\hline
\textbf{Condition} & \textbf{Recommended Paradigm} \\
\hline
\hline
Governing equations fully known; downstream tasks (e.g., control synthesis) require physically consistent Jacobians
& \textbf{DP} (hard-constrained): embeds exact physics as differentiable layers and enables efficient identification of unknown parameters via gradient-based optimization, yielding physically consistent Jacobians \\
\hline
Governing equations fully known; only parameter identification is required
& \textbf{DP or PINN}: DP (hard-constrained) efficiently learns unknown parameters within the known structure; PINN (soft-constrained) can also perform parameter identification but may require careful loss balancing and can exhibit slower convergence \\
\hline
Governing equations unavailable; abundant data available
& \textbf{NODE} (data-driven): learns dynamics directly from data with minimal structural assumptions \\
\hline
\end{tabular}
\end{table}
 
We acknowledge several limitations that suggest directions for future research. First, our benchmark is based on a single low-dimensional test system (SMIB); scalability to multi-machine power systems with higher-dimensional state spaces remains to be validated. Second, the noise model is limited to multiplicative Gaussian noise, whereas real-world power system measurements exhibit more complex characteristics, including bias, quantization, and non-stationary statistics. Third, while control performance is evaluated using LQR, extending the benchmark to advanced controllers such as Model Predictive Control would enable a more comprehensive assessment. Finally, 
the decision framework in Table~\ref{tab:decision_framework} covers only the two extremes of fully known and fully unknown governing equations. 
Since many practical problems fall between these extremes,
hybrid architectures that combine partial physical knowledge with data-driven components represent a promising direction for future work. One natural instance is a DP framework augmented with neural correction terms in the spirit of UDEs, which could retain the interpretability of DP while absorbing unmodeled dynamics through data-driven residuals.

\appendix
\section{Hyperparameter Settings}
\label{sec:hyperparams}

To ensure the reproducibility of the reported results, we provide the detailed hyperparameter configurations for the NODE, PINN, and DP frameworks. Table \ref{tab:hyperparams} summarizes the specific configurations used in our experiments.

\begin{table*}[h]
\caption{Hyperparameter configurations for PINN, NODE, and DP}
\label{tab:hyperparams}
\centering
\renewcommand{\arraystretch}{1.25} 
\setlength{\tabcolsep}{3pt}        
\begin{tabular}{l|l||c|c|c|c|c|c|c}
\hline
\textbf{Model} & \textbf{Parameter} 
    & Fig.~\ref{fig:smooth-trajectory}  
    & Fig.~\ref{fig:oscillation-trajectory} 
    & Table \ref{tab:l2err_forward_noise} 
    & Fig.~\ref{fig:oscillation-trajectory-noise} 
    & Fig.~\ref{fig:convergence_behavior}
    & Table \ref{tab:l2err_inverse}
    & Fig.~\ref{fig:lqr-node}\\
\hline
\hline
\multirow{4}{*}{\textbf{PINN }} 
 & Epochs & $10^5$ & $2\cdot 10^5$ & - & -& $2\cdot10^4$ & $2\cdot 10^4$ & - \\
 & Learning Rate & $10^{-3}$ & $10^{-3}$ & - & - & $10^{-3}$ & $10^{-3}$ & - \\
 & $\lambda_d$& $0$ & $0$ & - & - & $[0,0.1,0.5,1.0]$ & $1.0$ & - \\
 & $\lambda_i$& $2.0$ & $2.0$ & - & - & $2.0$ & $2.0$ & - \\
\hline
\multirow{3}{*}{\textbf{NODE }} 
 & Epochs & $10^3$ & $2\cdot 10^3$ & $[1000,2000,4000]$ & $4000$ & - & - & $2000$ \\
 & Learning Rate & $10^{-3}$ & $10^{-3}$ & $10^{-3}$ & $10^{-3}$ & -  & - & $10^{-3}$ \\
 & Step Length ($m$) & $10$ & $10$ & $[10,30, 60]$ & $60$ & - & - & $10$ \\
\hline
\multirow{3}{*}{\textbf{DP}} 
 & Epochs & - & - & - & - & $10^4$ & $4000$ & -\\
 & Learning Rate & - & - & - & - & $10^{-3}$ & $10^{-2}$ & - \\
 & Step Length ($m$) & - & - & - & - & $10$ & $[10,30,60]$ & - \\
\hline
\end{tabular}
\end{table*}

\printcredits

\bibliographystyle{cas-model2-names}
\bibliography{main-arxiv}

\end{document}